\documentclass[10pt,twocolumn,letterpaper]{article}

\usepackage[margin=0.75in]{geometry}
\usepackage{graphicx}
\usepackage{booktabs}
\usepackage{amsmath}
\usepackage{hyperref}
\usepackage{natbib}
\usepackage{caption}
\usepackage{subcaption}
\usepackage{float}
\usepackage{microtype}
\usepackage{times}
\usepackage{placeins}

\title{\textbf{Feature Dimensionality Outweighs Model Complexity in Breast Cancer Subtype Classification Using TCGA-BRCA Gene Expression Data}}

\author{
    Meena Al Hasani \\
    Independent Researcher
}

\date{}

\begin{document}

\maketitle

\begin{abstract}
Accurate classification of breast cancer subtypes from gene expression data is critical for diagnosis and treatment selection. However, such datasets are characterized by high dimensionality and limited sample size, posing challenges for machine learning models.

In this study, we evaluate the impact of model complexity and feature selection on subtype classification performance using TCGA-BRCA gene expression data. Logistic regression, random forest, and support vector machine (SVM) models were trained using varying numbers of highly variable genes (50 to 20,518). Performance was evaluated using stratified 5-fold cross-validation and assessed with accuracy and macro F1 score. While all models achieved high accuracy, macro F1 analysis revealed substantial differences in subtype-level performance. Logistic regression demonstrated the most stable and balanced performance across subtypes, including improved detection of rare classes. Random forest underperformed on minority subtypes despite strong overall accuracy, while SVM showed sensitivity to feature dimensionality. These findings highlight the importance of model simplicity, evaluation metrics, and feature selection in high-dimensional biological classification tasks.
\end{abstract}

\section{Introduction}

Breast cancer is a heterogeneous disease composed of multiple molecular subtypes, including Luminal A, Luminal B, HER2-enriched, Basal-like, and Normal-like \cite{sorlie2001}. These subtypes differ in prognosis and therapeutic response, making accurate classification essential for clinical decision-making. Standardized subtype classification was further enabled by the development of the PAM50 gene signature, a 50-gene panel that established a reproducible framework for molecular subtyping in clinical settings \cite{parker2009}.

Gene expression profiling enables subtype classification but presents unique challenges due to the high dimensionality of genomic data relative to sample size. This imbalance increases the risk of overfitting, particularly for complex machine learning models. While advanced models such as random forest and support vector machines are often assumed to outperform simpler models, their effectiveness in small-sample, high-dimensional biological settings remains unclear.

Additionally, traditional evaluation metrics such as accuracy may obscure poor performance on minority subtypes. Metrics such as macro F1 score, which equally weights all classes, provide a more informative assessment in imbalanced datasets. This study investigates:
\vspace{-0.2cm}
\begin{enumerate}
\setlength{\itemsep}{0pt}
\setlength{\parskip}{0pt}
\setlength{\parsep}{0pt}
    \item Whether increased model complexity improves classification performance
    \item How feature selection improves model behavior
    \item How evaluation metrics influence model interpretation
    \item Subtype-specific performance differences across models
\end{enumerate}

\FloatBarrier
\section{Related Work}

Machine learning has been increasingly applied to breast cancer subtype classification using gene expression data, building on early work demonstrating the feasibility of cancer classification from gene expression profiles \cite{golub1999, statnikov2005}. More broadly, machine learning has become a central tool in genomics for analyzing high-dimensional biological data \cite{libbrecht2015}. Yu et al. \cite{yu2020} applied multiple machine learning models to RNA-seq data from TCGA to classify breast cancer into the five intrinsic molecular subtypes, demonstrating the feasibility of automated subtype identification from high-dimensional genomic profiles. Similarly, Wu and Hicks \cite{wu2021} evaluated support vector machines, K-nearest neighbor, Naive Bayes, and decision tree classifiers on TCGA gene expression data, finding that SVM outperformed competing models. However, both studies primarily rely on accuracy as the evaluation metric, which may obscure poor performance on minority subtypes in imbalanced datasets.

Deep learning approaches have also been explored for this problem, reflecting broader trends in applying deep learning to healthcare and biomedical data \cite{esteva2019}. Chhikara et al. \cite{chhikara2021} proposed a two-stage framework combining an autoencoder for dimensionality reduction with a deep neural network classifier, reducing the feature space from over 20,000 genes to 500 and achieving strong classification accuracy on TCGA data. While this work demonstrates the value of dimensionality reduction as a preprocessing step, it does not systematically evaluate how different levels of feature reduction affect model behavior across multiple classifier types, nor does it report subtype-level performance metrics that reveal minority class failure modes.

Feature selection is widely recognized as a critical step in machine learning workflows for high-dimensional omics data. Sanz et al. \cite{sanz2017} demonstrated that removing redundant and irrelevant genes improves model generalization and reduces overfitting in high-dimensional biological classification tasks. Despite this, existing studies on breast cancer subtype classification rarely investigate the relationship between feature dimensionality and classifier performance in a systematic and comparative manner.

A consistent limitation across prior work is the reliance on aggregate metrics such as accuracy, which obscures subtype-specific performance differences. In datasets with significant class imbalance such as TCGA-BRCA, where Luminal A samples constitute over half the cohort, high accuracy can be achieved by models that effectively ignore minority subtypes. This study addresses these gaps by systematically evaluating three classifiers of increasing complexity across five feature dimensionality levels, using macro F1 score as the primary evaluation metric and reporting per-subtype performance to expose failure modes invisible to accuracy-based evaluation.

\FloatBarrier
\section{Dataset}

Gene expression data and subtype annotations were obtained from the TCGA-BRCA dataset \cite{weinstein2013}. After aligning expression and clinical data by patient ID and removing missing subtype labels, a total of 981 samples were retained across five subtypes, as shown in Table~\ref{tab:dataset}. Gene expression features were derived from RNA-seq data, where each feature represents the expression level of a gene, resulting in 20,518 features per sample.

The dataset presents several challenges for machine learning classification. It is high-dimensional relative to the number of samples, increasing the risk of overfitting. Additionally, the subtype distribution is imbalanced, with BRCA\_LumA comprising the majority of samples and BRCA\_Normal representing a small minority. These characteristics motivate the use of robust evaluation metrics and cross-validation strategies.

\begin{table}[H]
    \centering
    \caption{Detailed distribution of subtypes from the TCGA-BRCA dataset.}
    \label{tab:dataset}
    \begin{tabular}{lc}
        \toprule
        \textbf{Subtype} & \textbf{Sample Count} \\
        \midrule
        BRCA\_Basal  & 171 \\
        BRCA\_Her2   & 78 \\
        BRCA\_LumA   & 499 \\
        BRCA\_LumB   & 197 \\
        BRCA\_Normal & 36 \\
        \midrule
        Total        & 981 \\
        \bottomrule
    \end{tabular}
\end{table}

\FloatBarrier
\section{Evaluation Metrics}

Model performance was evaluated using accuracy, macro F1 score, weighted F1 score, and per-subtype F1 scores. These metrics provide complementary perspectives on model performance, particularly in the presence of class imbalance.

Accuracy measures the proportion of correctly classified samples. While it provides a general measure of performance, it can be misleading in imbalanced datasets where dominant classes disproportionately influence the score, a well-documented issue in machine learning \cite{he2009}.

Macro F1 score computes the F1 score independently for each class and then averages across all classes, assigning equal weight to each subtype. This makes it more suitable for evaluating performance in imbalanced datasets. Prior work has shown that metric choice significantly affects the interpretation of classification performance, particularly in imbalanced settings \cite{sokolova2009}. This metric is therefore used as the primary evaluation metric in this study.

Weighted F1 score accounts for class imbalance by weighting each class according to its number of samples. While it provides a compromise between accuracy and macro F1, it still tends to favor dominant classes.

Per-subtype F1 scores were analyzed to assess model performance on individual breast cancer subtypes. This enables identification of subtype-specific weaknesses that may not be apparent from aggregate metrics.

\subsection{Accuracy}

Accuracy measures the overall proportion of correctly classified samples across all subtypes. In imbalanced datasets such as TCGA-BRCA, accuracy is highly influenced by dominant classes such as BRCA\_LumA. As a result, models can achieve high accuracy even when performing poorly on minority subtypes.

\begin{equation}
    \text{Accuracy} = \frac{TP + TN}{TP + TN + FP + FN}
\end{equation}

\subsection{Macro F1-Score}

Macro F1 score evaluates model performance by computing the F1 score for each subtype independently and averaging across all classes. This ensures that each class contributes equally, regardless of its frequency. As a result, Macro F1 provides a more reliable measure of performance in imbalanced datasets where minority subtypes are clinically important.

\begin{equation}
    \text{Macro F1} = \frac{1}{C} \sum_{c=1}^{C} \frac{2 \cdot P_c \cdot R_c}{P_c + R_c}
\end{equation}

\subsection{Weighted F1-Score}

Weighted F1 score computes the average F1 score across all classes, weighted by the number of samples in each class. This provides a balance between overall accuracy and class-level performance but still biases results toward dominant subtypes.

\subsection{Per-Subtype F1-Score}

Per-subtype F1 scores directly evaluate performance on each individual breast cancer subtype. This enables detailed analysis of model behavior across classes and highlights weaknesses in minority subtype classification.

\subsection{Metric Selection}

Given the strong imbalance in TCGA-BRCA, macro F1 score was selected as the primary evaluation metric. Accuracy and weighted F1 were used as supporting metrics, while per-subtype F1 scores provide additional insight into model behavior across individual subtypes.

\FloatBarrier
\section{Baseline Evaluation}

A majority class baseline was used to provide a reference point for model performance. This baseline predicts all samples as BRCA\_LumA, the dominant subtype in the dataset (499 out of 981 samples). While this approach yields relatively high accuracy, it fails to capture meaningful subtype distinctions and performs poorly under macro F1 evaluation. All models were evaluated using stratified 5-fold cross-validation to preserve subtype distribution, with mean and standard deviation reported across folds.

\subsection{Majority Class Baseline}

The majority class baseline highlights the limitations of accuracy as a standalone metric in imbalanced datasets. Although it achieves an accuracy of 0.509, its macro F1 score is near zero, reflecting complete failure to classify minority subtypes.

\subsection{Evaluation and Results}

As shown in Table~\ref{tab:baseline}, all machine learning models substantially outperform the majority class baseline. Logistic regression achieves the highest overall performance, with both strong accuracy and macro F1 score, indicating balanced classification across subtypes. Random forest achieves comparable accuracy but lower macro F1, suggesting bias toward dominant classes. SVM demonstrates lower overall performance, particularly under macro F1 evaluation.

\begin{table}[H]
    \centering
    \caption{Baseline and model performance at full feature set (20,518 genes).}
    \label{tab:baseline}
    \begin{tabular}{lcc}
        \toprule
        \textbf{Model} & \textbf{Accuracy} & \textbf{Macro F1} \\
        \midrule
        Majority Class Baseline & 0.509 & 0.140 \\
        Logistic Regression     & 0.861 & 0.795 \\
        Random Forest           & 0.850 & 0.689 \\
        SVM                     & 0.691 & 0.587 \\
        \bottomrule
    \end{tabular}
\end{table}

\subsection{Performance Across Feature Sizes}

Accuracy increased with the number of genes for logistic regression and random forest, with both models achieving peak performance at 1000--20,000 genes ($\sim$0.85--0.86). In contrast, SVM performance peaked at intermediate feature sizes (around 1000 genes) and declined at higher dimensionality, indicating sensitivity to large feature spaces.

\subsection{Macro F1 Reveals Model Differences}

Macro F1 analysis revealed substantial differences not captured by accuracy alone. Logistic regression consistently achieved the highest macro F1 scores ($\sim$0.79--0.80), indicating balanced performance across subtypes. Random forest underperformed ($\sim$0.69--0.70), suggesting bias toward dominant classes. SVM showed competitive performance at moderate feature sizes but degraded at high dimensionality.

\subsection{Subtype-Specific Performance}

Subtype-level analysis revealed substantial variation in model behavior across classes. Logistic regression maintained consistent performance across all subtypes, including minority classes (BRCA\_Normal F1 $\approx$ 0.62). Random forest achieved strong performance on dominant subtypes but performed poorly on rare classes (BRCA\_Normal F1 $\approx$ 0.21). SVM showed improved performance on certain subtypes such as LumB and Normal but lacked overall stability. Notably, initial single train-test split results suggested that logistic regression failed on the BRCA\_Normal subtype. However, cross-validation revealed that this was an artifact of data splitting, with true performance substantially higher. This highlights the importance of robust evaluation strategies in high-dimensional, imbalanced datasets.

\FloatBarrier
\section{Main Approach}

This study evaluates the interaction between feature dimensionality and model complexity in breast cancer subtype classification using gene expression data. The workflow consists of preprocessing, variance-based feature selection, model training, and cross-validated evaluation across multiple feature sizes.

Three models of increasing complexity were compared: logistic regression (linear), random forest (ensemble), and support vector machine with an RBF kernel (nonlinear). This design enables direct comparison of how model complexity interacts with high-dimensional feature spaces.

\subsection{Preprocessing}

Samples with missing subtype labels were removed. Gene expression features were standardized using z-score normalization for models sensitive to feature scaling (logistic regression and SVM), while no scaling was applied to random forest.

\subsection{Feature Selection}

To address the high dimensionality of gene expression data, features were selected based on variance within the training data. Variance-based feature selection is a widely used heuristic in genomic studies, operating under the assumption that genes exhibiting greater expression variability across samples are more likely to carry subtype-discriminative signal \cite{sanz2017, guyon2003}. Feature selection is particularly important in bioinformatics applications, where datasets often contain thousands of genes but relatively few samples \cite{saeys2007}. While this approach does not guarantee selection of the most biologically meaningful features, it provides a computationally efficient and parameter-free method for dimensionality reduction that avoids data leakage when applied within each training fold.

For each experiment, genes were ranked by variance and the top $N$ genes were selected, where $N \in \{50, 75, 100, 1000, 20518\}$. This range allows for systematic evaluation of model performance under progressively reduced feature spaces, from aggressive dimensionality reduction to the full gene set. Importantly, feature selection was performed independently within each training fold during cross-validation to prevent data leakage and preserve the integrity of performance estimates.

To assess the biological relevance of the selected features, genes consistently selected across all five cross-validation folds at the 1,000-gene threshold were examined. Of the 885 stably selected genes, several are established breast cancer subtype markers, including ERBB2, which defines the HER2-enriched subtype, KRT8, KRT14, and KRT19, which distinguish luminal from basal subtypes, and SCGB2A2 and XBP1, which are associated with luminal breast cancer. This suggests that variance-based selection captures genuine subtype-discriminative signal rather than arbitrary biological or technical variation. However, this analysis serves as a sanity check rather than a formal biomarker discovery, and the presence of known markers does not preclude the inclusion of non-subtype-related signal in the selected feature set.

\subsection{Models}

Three models of increasing complexity were evaluated to assess how model capacity interacts with feature dimensionality in high-dimensional biological data.

\subsubsection{Logistic Regression}

Logistic regression serves as a linear baseline model. It assumes a linear decision boundary in feature space and is well-suited for high-dimensional datasets due to its use of L2 regularization (C = 1.0, default), which helps prevent overfitting. Class weights were balanced to account for subtype imbalance.

\subsubsection{Random Forest}

Random forest is an ensemble learning method that constructs multiple decision trees using bootstrap sampling and feature subsampling \cite{breiman2001}. It is capable of capturing nonlinear relationships and interactions between genes. A total of 500 estimators were used with balanced class weights. However, in imbalanced datasets, it may preferentially model dominant classes, leading to reduced performance on minority subtypes.

\subsubsection{Support Vector Machine}

A support vector machine with a radial basis function (RBF) kernel was used to model nonlinear decision boundaries \cite{cortes1995} (C = 1.0, gamma = \texttt{scale}, default). The RBF kernel enables the model to capture complex relationships between genes. However, SVMs are sensitive to high-dimensional feature spaces and limited sample sizes, which can lead to reduced stability and performance when using very large feature sets.

\subsection{Cross-Validation Strategy}

Model performance was evaluated using stratified 5-fold cross-validation to preserve subtype distribution. Within each fold, feature selection was performed using only the training data before applying the model to the validation fold. This ensures that performance estimates reflect true generalization rather than benefiting from information leakage. To assess whether observed performance differences between models were statistically significant, pairwise Wilcoxon signed-rank tests were applied to per-fold macro F1 scores at the 1,000-gene feature size.

\subsection{Performance Summary}

Table~\ref{tab:results} summarizes cross-validated performance across feature sizes. Logistic regression consistently achieves the highest macro F1 scores, indicating balanced performance across subtypes. Random forest attains comparable accuracy but lower macro F1, reflecting bias toward dominant classes. SVM performs competitively at intermediate feature sizes but degrades at high dimensionality.

\FloatBarrier
\begin{table}[H]
    \centering
    \caption{Cross-validated model performance across feature sizes.}
    \label{tab:results}
    \begin{tabular}{llccc}
        \toprule
        \textbf{Genes} & \textbf{Model} & \textbf{Acc} & \textbf{Macro F1} & \textbf{Wtd F1} \\
        \midrule
        20518 & LR  & 0.861 & 0.795 & 0.860 \\
              & RF  & 0.850 & 0.689 & 0.831 \\
              & SVM & 0.691 & 0.587 & 0.693 \\
        \midrule
        1000  & LR  & 0.838 & 0.789 & 0.838 \\
              & RF  & 0.833 & 0.700 & 0.812 \\
              & SVM & 0.804 & 0.736 & 0.806 \\
        \midrule
        100   & LR  & 0.777 & 0.716 & 0.782 \\
              & RF  & 0.808 & 0.662 & 0.784 \\
              & SVM & 0.766 & 0.677 & 0.768 \\
        \midrule
        75    & LR  & 0.775 & 0.720 & 0.781 \\
              & RF  & 0.811 & 0.657 & 0.787 \\
              & SVM & 0.769 & 0.678 & 0.771 \\
        \midrule
        50    & LR  & 0.721 & 0.656 & 0.734 \\
              & RF  & 0.790 & 0.608 & 0.758 \\
              & SVM & 0.740 & 0.658 & 0.743 \\
        \bottomrule
    \end{tabular}
\end{table}

\section{Discussion}

The results demonstrate that increasing model complexity does not necessarily improve performance in high-dimensional, small-sample biological datasets. Logistic regression consistently achieved the strongest overall performance in terms of macro F1 score, indicating balanced classification across subtypes. Random forest achieved comparable accuracy but lower macro F1, suggesting that its performance was driven by dominant classes. SVM demonstrated competitive performance at intermediate feature sizes but showed reduced stability at higher dimensionality. These findings highlight the importance of jointly evaluating feature dimensionality, model complexity, and evaluation metrics rather than optimizing for accuracy alone.

Pairwise Wilcoxon signed-rank tests were conducted to assess the statistical significance of performance differences between models. Logistic regression achieved higher macro F1 scores than both random forest and SVM across cross-validation folds, however neither comparison reached statistical significance at the 0.05 threshold (LR vs RF: $p = 0.0625$, LR vs SVM: $p = 0.0625$, RF vs SVM: $p = 1.000$). This is attributable to the limited statistical power of 5-fold cross-validation, where the minimum achievable $p$-value is $0.0625$ with five paired observations. The performance differences should therefore be interpreted as consistent practical trends rather than formally significant results.

\vspace{-0.3cm}
\subsection{Effect of Feature Reduction Across Models}

\begin{figure}[H]
    \centering
    \includegraphics[width=\columnwidth]{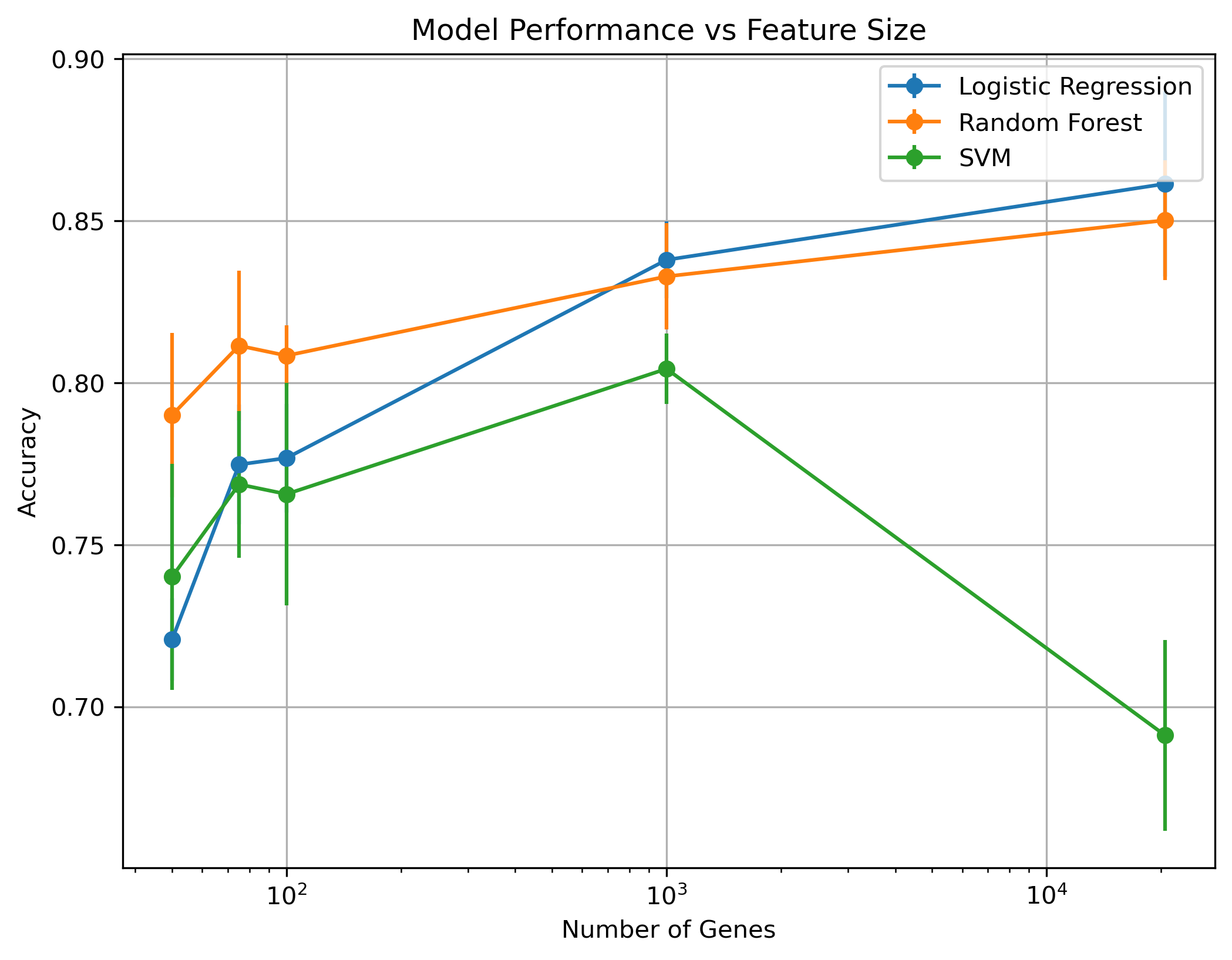}
    \caption{Model accuracy versus number of genes used for feature selection. Error bars indicate standard deviation across 5-fold cross-validation.}
    \label{fig:accuracy}
\end{figure}

\begin{figure}[H]
    \centering
    \includegraphics[width=\columnwidth]{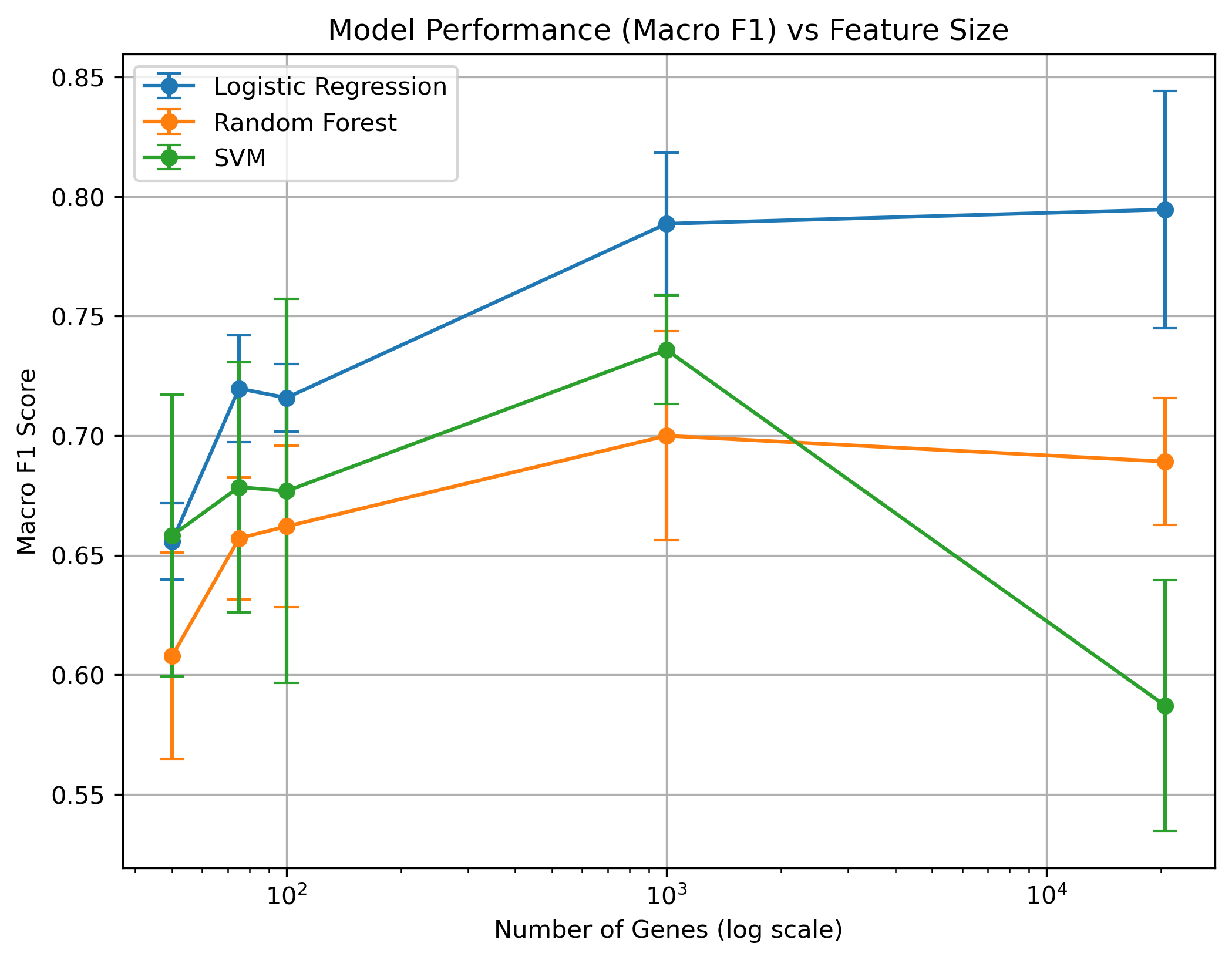}
    \caption{Macro F1 score versus number of genes. Unlike accuracy, macro F1 equally weights all subtypes and exposes performance differences obscured by class imbalance.}
    \label{fig:macrof1}
\end{figure}

Feature dimensionality had a significant impact on model performance. All models improved as the number of genes increased from 50 to 1000, suggesting that important subtype-discriminative information is captured within a relatively small subset of highly variable genes. However, performance gains plateaued beyond approximately 1000 genes, indicating diminishing returns from additional features. This trend is consistent with Figures~\ref{fig:accuracy} and \ref{fig:macrof1}, where performance stabilizes beyond approximately 1000 genes. Notably, SVM performance declined at higher dimensionality, suggesting sensitivity to high-dimensional feature spaces and limited sample size.

\begin{figure}[H]
    \centering
    \includegraphics[width=\columnwidth, height=0.51\textheight, keepaspectratio]{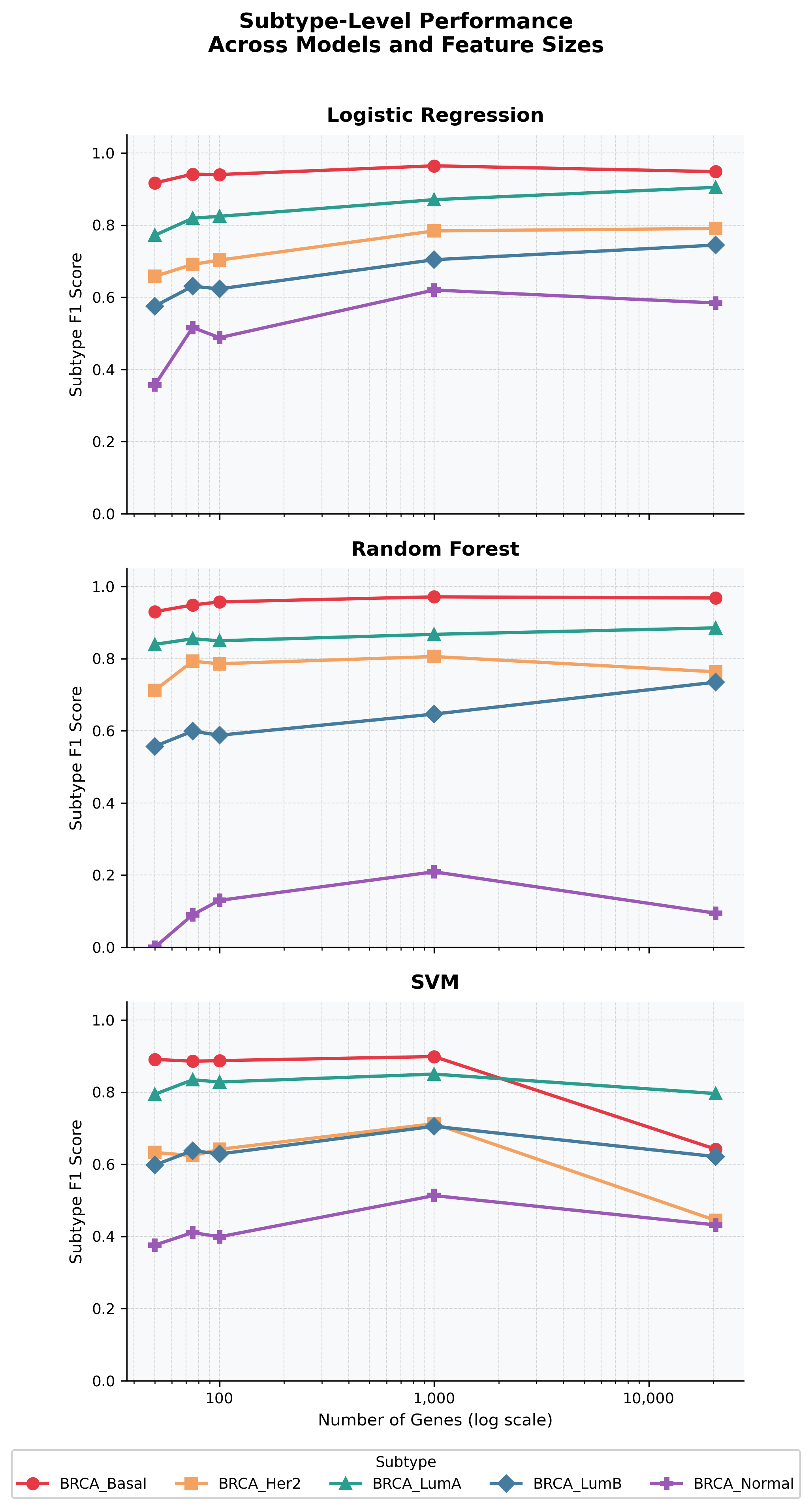}
    \caption{Per-subtype F1 scores across feature sizes for logistic regression, random forest, and SVM. Each line represents one breast cancer subtype. BRCA\_Basal achieves consistently high F1 across all models while BRCA\_Normal shows the greatest variability.}
    \label{fig:subtype}
\end{figure}

\subsection{Subtype-Specific Performance}

Subtype-level analysis revealed that classification performance varies significantly across breast cancer subtypes. BRCA\_Basal and BRCA\_LumA were consistently classified with high F1 scores across all models, indicating strong separability in gene expression space. In contrast, BRCA\_LumB and BRCA\_Her2 showed moderate performance, suggesting that these subtypes may share overlapping gene expression patterns, making them more difficult to separate using standard machine learning models.

The BRCA\_Normal subtype exhibited the greatest variability, reflecting both its small sample size and biological similarity to other subtypes. Logistic regression achieved the most consistent performance across subtypes, while random forest struggled with minority classes. SVM demonstrated moderate improvement for certain subtypes but lacked overall consistency.

\begin{figure}[H]
    \centering
    \includegraphics[width=\columnwidth]{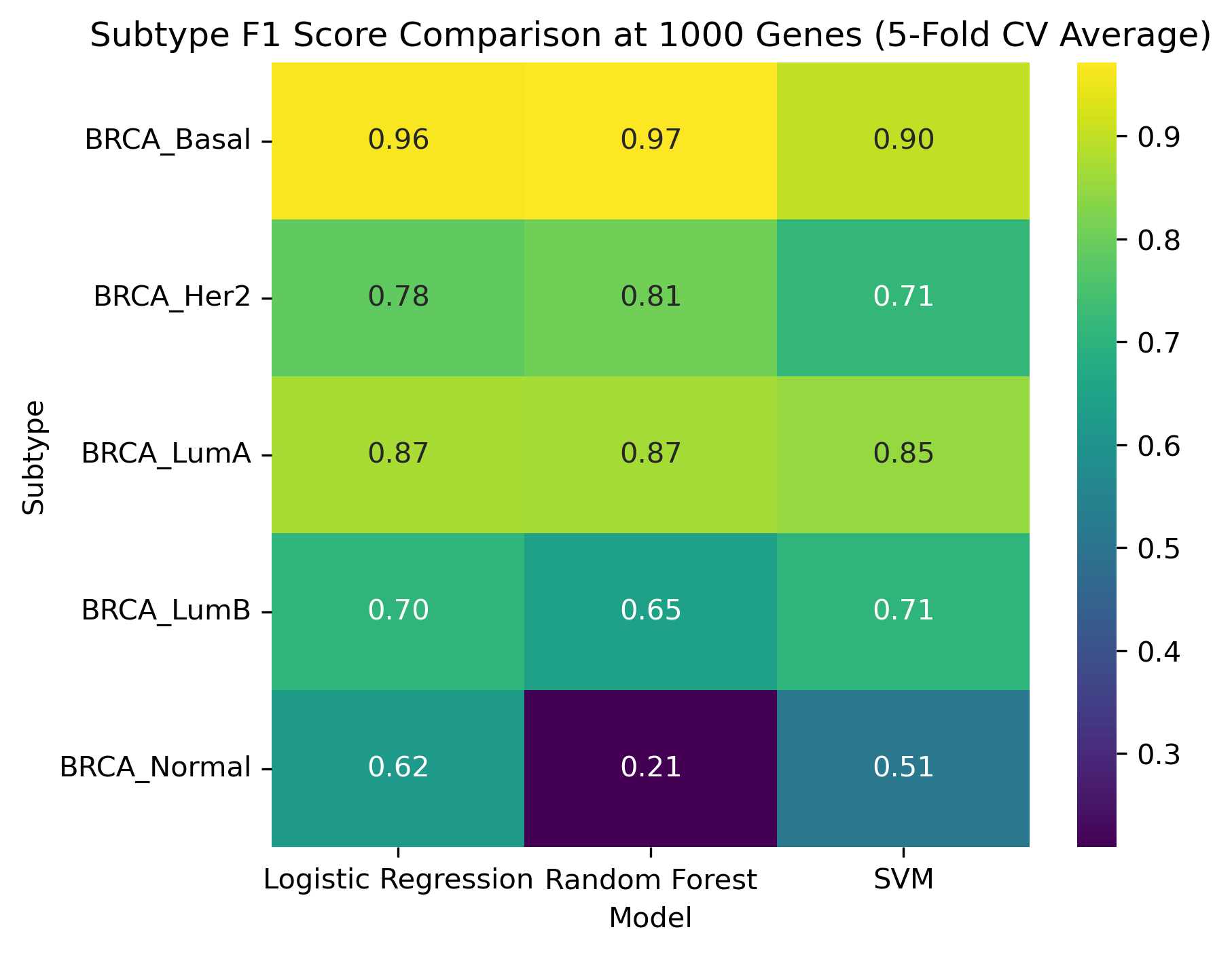}
    \caption{Per-subtype F1 scores at 1,000 genes averaged across 5-fold cross-validation. Random forest achieves the lowest F1 on BRCA\_Normal (0.21) while logistic regression maintains the most consistent performance across all subtypes.}
    \label{fig:heatmap}
\end{figure}

\subsection{Error Analysis}

Model errors were primarily concentrated in minority and biologically overlapping subtypes. The BRCA\_Normal subtype, which has the smallest number of samples, exhibited the most variability in classification performance across models. Random forest consistently underperformed on this class, while logistic regression and SVM showed moderate improvement.

The BRCA\_LumB subtype also demonstrated reduced performance compared to BRCA\_LumA, suggesting partial overlap in gene expression patterns between these subtypes. In contrast, BRCA\_Basal was consistently classified with high performance across all models, indicating strong separability in gene expression space. These results indicate that classification errors are driven by both class imbalance and intrinsic biological similarity between subtypes, reinforcing the need for evaluation metrics that capture subtype-level performance.

\subsection{Model Recommendation}

Based on the results of this study, logistic regression with L2 regularization trained on the top 1,000 highest-variance genes represented the strongest overall configuration evaluated in this study for breast cancer subtype classification using TCGA-BRCA RNA-seq data. This configuration consistently achieved the highest macro F1 score across all feature sizes, demonstrated the most balanced per-subtype performance including minority classes, and showed the greatest robustness to class imbalance. Its computational efficiency and interpretability further support its suitability for high-dimensional biological classification tasks.

Random forest is not recommended as a primary classifier in this setting. Despite achieving comparable overall accuracy, it systematically underperformed on minority subtypes, particularly BRCA\_Normal, suggesting that its ensemble mechanism favors dominant classes in imbalanced datasets. SVM demonstrated competitive performance at intermediate feature sizes but showed significant degradation at full dimensionality, making it sensitive to the choice of feature set and less reliable across varying experimental conditions.

\FloatBarrier
\section{Limitations}

Several limitations of this study should be acknowledged. First, variance-based feature selection is a heuristic that does not distinguish between subtype-discriminative signal and other sources of biological or technical variation, such as patient age, tumor stage, or batch effects. Genes selected by variance may capture non-subtype-related variation, and future work incorporating pathway-based or supervised feature selection methods may improve biological interpretability.

Second, all models were evaluated using default hyperparameters without systematic tuning. A grid search or Bayesian optimization may alter the relative performance between models, particularly for SVM where kernel parameters can substantially affect behavior in high-dimensional spaces.

Third, this study relies on a single dataset, TCGA-BRCA, and results may not generalize to other cohorts with different patient populations, sequencing protocols, or sample collection methods. External validation on an independent dataset such as METABRIC would strengthen the generalizability of these findings.

Fourth, the BRCA\_Normal subtype is represented by only 36 samples, making performance estimates for this class less reliable than for larger subtypes regardless of cross-validation strategy.

Finally, pairwise Wilcoxon signed-rank tests did not reach statistical significance at the 0.05 threshold (LR vs RF: $p = 0.0625$, LR vs SVM: $p = 0.0625$), likely due to the limited statistical power of 5-fold cross-validation. The observed performance differences should be interpreted as practically meaningful trends rather than formally validated conclusions. Future work using repeated cross-validation or larger datasets may provide sufficient power for formal significance testing.

\FloatBarrier
\section{Conclusion}

This study demonstrates that, in gene expression-based breast cancer subtype classification, simpler models such as logistic regression can outperform more complex models in terms of balanced performance. The results show that accuracy alone is insufficient for evaluation, and that macro F1 provides a more reliable metric in the presence of class imbalance. Additionally, cross-validation is essential to avoid misleading conclusions, particularly for minority subtypes, and variance-based feature selection improves model stability without sacrificing performance. These findings highlight practical considerations for applying machine learning to high-dimensional biomedical data and emphasize the importance of robust evaluation strategies.

\FloatBarrier
\section{Future Work}

Future work could explore the use of larger datasets or multi-cohort integration to improve generalization. Incorporating additional data types, such as methylation or proteomics, may further enhance subtype classification performance.

More advanced techniques for handling class imbalance, such as synthetic oversampling or cost-sensitive learning, could improve performance on minority subtypes. Additionally, exploring alternative feature selection methods beyond variance, such as pathway-based approaches, may provide more biologically interpretable results.

Finally, extending this framework to deep learning models could further evaluate the trade-off between model complexity and data limitations.

\section*{Author Contributions}

The author conceived the study, implemented the models, performed the data analysis, and wrote the manuscript.

{\small
\bibliographystyle{plain}
\bibliography{references}
}

\end{document}